\documentclass[10pt,twocolumn,letterpaper]{article}

\usepackage{cvpr}              
\usepackage{tabularx}
\usepackage{booktabs}
\newcolumntype{C}{>{\centering\arraybackslash}X}
\usepackage[table]{xcolor}
\usepackage{bbding}
\newcommand\blfootnote[1]{%
  \begingroup
  \renewcommand\thefootnote{}\footnote{#1}%
  \addtocounter{footnote}{-1}%
  \endgroup
}
\usepackage[accsupp]{axessibility}  


\usepackage{algorithm}
\usepackage{algorithmic}
\usepackage{multirow}
\usepackage[table]{xcolor}

\usepackage{amsmath}
\DeclareMathOperator*{\argmin}{arg\,min}
\DeclareMathOperator*{\argmax}{arg\,max}

\definecolor{cvprblue}{rgb}{0.21,0.49,0.74}
\usepackage[pagebackref,breaklinks,colorlinks,allcolors=cvprblue]{hyperref}

\def\paperID{NA} 
\def\confName{CVPR}
\def\confYear{2026}

\title{Adaptive Action Chunking at Inference-time for Vision-Language-Action Models}

\author{
    Yuanchang Liang$^{1}$ \quad
    Xiaobo Wang$^{2,3}$\textsuperscript{\Envelope} \quad
    Kai Wang$^{1}$ \quad
    Shuo Wang$^{4}$ \\[1.5mm] 
    Xiaojiang Peng$^{5}$ \quad
    Haoyu Chen$^{6}$ \quad
    David Kim Huat Chua$^{1}$ \quad
    Prahlad Vadakkepat$^{1}$ \\[3mm] 
    \small 
    $^{1}$National University of Singapore \quad
    $^{2}$Shenzhen University of Advanced Technology \\[1mm] 
    \small 
    $^{3}$Sangfor Technologies Inc. \quad
    $^{4}$Mininglamp Technology \quad
    $^{5}$Shenzhen Technology University \quad
    $^{6}$City University of Hong Kong \\[2mm]
    \small
    \textit{liangyuanchang@u.nus.edu, wangxiaobo@suat-sz.edu.cn}
}

\begin{document}
\maketitle
\blfootnote{\textsuperscript{\Envelope} Corresponding author. This work was supported by the Innovative Team Project in Guangdong Province under Grant 2025KCXTD040.}

\begin{abstract}

In Vision-Language-Action (VLA) models, action chunking (\textit{i.e.}, executing a sequence of actions without intermediate replanning) is a key technique to improve robotic manipulation abilities. However, a large chunk size reduces the model's responsiveness to new information, while a small one increases the likelihood of mode-jumping, jerky behavior resulting from discontinuities between chunks. Therefore, selecting the optimal chunk size is an urgent demand to balance the model's reactivity and consistency. Unfortunately, a dominant trend in current VLA models is an empirical fixed chunk length at inference-time, hindering their superiority and scalability across diverse manipulation tasks. To address this issue, we propose a novel Adaptive Action Chunking (AAC) strategy, which exploits \textbf{action entropy} as the cue to adaptively determine the chunk size based on current predictions. Extensive experiments on a wide range of simulated and real-world robotic manipulation tasks have demonstrated that our approach substantially improves performance over the state-of-the-art alternatives. The videos and source code are publicly available at \url{https://lance-lot.github.io/adaptive-chunking.github.io/}.

\end{abstract}    
\section{Introduction}
\label{sec:intro}
\begin{figure}[t]
\centering
\includegraphics[width=\linewidth]{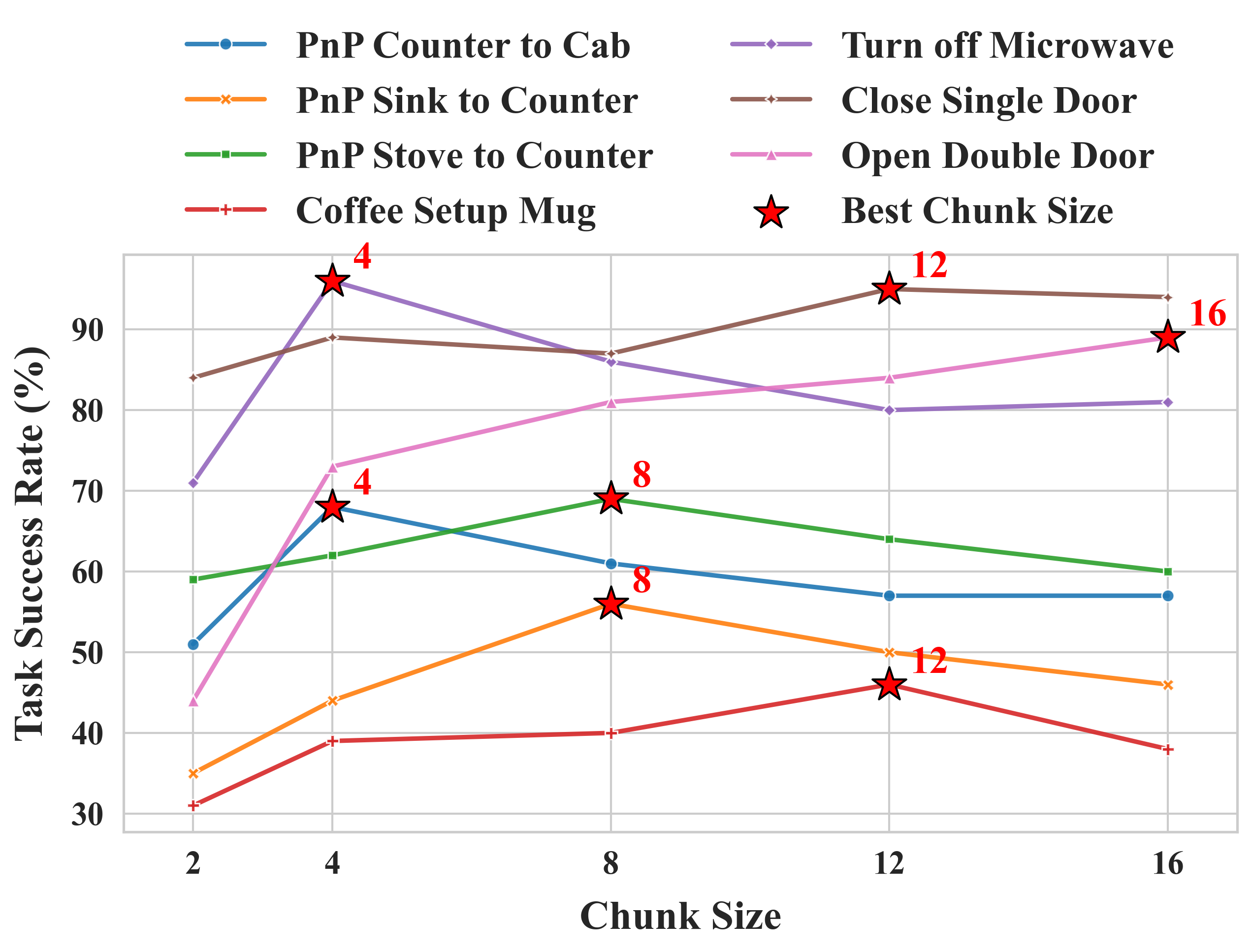}
\caption{\textbf{Effects of action chunk sizes}. At inference-time, the success rates of the GR00T N1.5~\cite{bjorck2025gr00t} on different tasks of RoboCasa Kitchen~\cite{nasiriany2024robocasa} are highly related to the action chunk size. It can be observed that it is difficult and sub-optimal to empirically set a fixed value for various manipulation tasks.
}
\label{fig:replan}
\end{figure}

Vision-Language-Action (VLA) models have shown great potential to improve robotic manipulation abilities by generating robotic control actions from a variety of data modalities \cite{ma2024survey,bjorck2025gr00t,black2410pi0,shukor2025smolvla,yu2025forcevla,ze20243d,huang2025tactile, wang2025vla, li2025bridgevla, liu2025hybridvla, jiang2025rynnvla, shah2025learning}. Usually, they map basic information such as vision, language instructions, and robotic proprioception states, or additional information like tactile~\cite{huang2025tactile, bi2025vla}, force~\cite{yu2025forcevla}, and 3D information~\cite{yuan2025depthvla, zhen20243d, qu2025spatialvla, li2026pointvla, zhang20254d, singh2025og} into executable actions. Beyond that, earlier studies in imitation learning, such as ACT~\cite{zhao2023learning} and Diffusion Policy \cite{chi2023diffusion} have shown that executing a sequence of actions at each observation point, known as \textit{action chunking}, can effectively improve the success rates as it reduces compounding errors. Therefore, the action chunking technique is employed as a standard paradigm for VLA models. Jointly, the action chunk is simply set as a fixed length by empirical experiments. For instance, in GR00T N1.5~\cite{bjorck2025gr00t}, the chunk size is set as 16 at inference-time for the RoboCasa Kitchen~\cite{nasiriany2024robocasa} benchmark. $\pi_0$~\cite{black2410pi0} sets the inference chunk size as 16 or 25 for different benchmarks. Differently, SmolVLA~\cite{shukor2025smolvla} suggests a chunk size of 10 for the LIBERO~\cite{liu2023libero} benchmark. In summary, determining a proper chunk size is non-trivial as it is not consistent for different policies. Moreover, it is intuitive that the best chunk size for different manipulation tasks is also different. To demonstrate this, we performed experiments with GR00T N1.5~\cite{bjorck2025gr00t} across multiple tasks in RoboCasa Kitchen~\cite{nasiriany2024robocasa}.
As shown in~\cref{fig:replan}, the success rates of different tasks exhibit a strong dependency on the action chunk size.

To mitigate this dilemma at inference-time, ACT~\cite{zhao2023learning} integrates current and past predicted actions using Exponential Moving Average (EMA) to enhance temporal consistency. BID~\cite{liu2025bidirectional} employs a search strategy to select the optimal full action chunk from multiple candidates, where backward coherence and forward contrast scores are used to balance the policy's consistency and reactivity. TV-BID~\cite{khantest} further extends BID~\cite{liu2025bidirectional} by adapting the backward coherence and forward contrast scores according to the Total Variation Distance (TVD) between consecutive action chunk distributions. SGAC~\cite{so2025improving} selectively updates action chunks by comparing the cosine similarity between the first action of the previous planned chunk and the newly generated action. Nevertheless, all these methods set a fixed chunk size throughout a whole episode, relying on extensive empirical tuning for the action chunk size selection. Recent studies \cite{gou2024learning,chen2025adaptive} have begun to adaptively select the chunk size by reinforcement learning. Gou et al.~\cite{gou2024learning} determine whether to choose a single action or an entire action chunk based on a regret value function. However, this handcrafted value function fails to generalize across various tasks. Chen et al.~\cite{chen2025adaptive} introduce a learnable module that predicts chunk size, but this approach depends on task-specific reward signals available only in simulation, exhibiting limited applicability in real-world deployment. Consequently, adaptively selecting the optimal action chunk size at inference-time remains a challenging problem in practice.


In this paper, we propose a simple yet effective Adaptive Action Chunking (AAC) strategy, which exploits \textit{action entropy} as the cue to determine the optimal chunk size at each inference timestep dynamically. Specifically, as action entropy reflects the uncertainty level of the predicted action, a lower entropy implies higher action reliability during task execution. Inspired by this, AAC tries to adaptively determine the optimal action chunk size by maintaining a lower action entropy. In detail, we compute the maximum differential point of average action entropy across different chunk sizes meanwhile ensuring minimum action magnitude. In high-entropy cases, where the predicted actions are of low quality, AAC executes smaller chunks and replans more frequently to improve the model's reactivity. Conversely, when the entropy is low, longer action chunks are executed to enhance temporal consistency and inference efficiency. Notably, AAC performs action entropy computation and chunk-size adaptation only at inference-time, without additional training or architectural modifications. Therefore, it ensures the robustness and scalability of our method to various manipulation tasks. To sum up, the main contributions of this paper can be summarized as follows:
\begin{itemize}
    \item We analyze existing action chunking methods and highlight the importance of adaptive action chunking at inference-time for diffusion-based VLA models.  
    \item We propose AAC, a simple yet effective algorithm that exploits action entropy as the cue to dynamically select the action chunk size at inference-time.
    \item We conduct extensive experiments on a wide range of simulated and real-world robotic manipulation benchmarks, demonstrating the superiority of our approach over the state-of-the-art alternatives. 
\end{itemize}

\section{Related work}
\label{sec:related work}

\subsection{Vision-Language-Action models}
Inspired by the success of large pre-trained vision-language models (VLMs), Vision-Language-Action (VLA) models have shown great promise in robotic policy learning by leveraging diverse, robot-specific datasets. One line of VLA models~\cite{kim2025openvla, pertsch2025fast, brohan2023rt, zitkovich2023rt, cen2025worldvla, song2025accelerating} discretizes the continuous action space and predicts actions as tokens, showing that pre-trained VLMs can be effectively adapted to low-level action predictions. In contrast, a growing body of studies~\cite{black2410pi0, bjorck2025gr00t, shukor2025smolvla, li2025cogvla, zhao2025cot, kim2025fine, zheng2025x, chen2025fast, intelligence2025pi, song2025hume, koo2025hamlet, zhong2025dexgraspvla, hou2025dita} leverage action experts to predict continuous actions via diffusion or flow-matching, enabling more accurate predictions. For instance, GR00T N1.5~\cite{bjorck2025gr00t} conditions a flow-matching head on VLM features to generate action chunks. CogVLA~\cite{li2025cogvla} systematically compares action modules and finds that diffusion action transformers scale favorably when conditioned on VLM representations. CoT-VLA~\cite{zhao2025cot} incorporates explicit visual chain-of-thought (CoT) reasoning into VLM by autoregressively predicting future image frames as visual goals. In this paper, we propose an adaptive action chunking strategy that operates at inference-time for VLA models, without requiring additional training or architectural modifications. Therefore, without loss of generality, this work adopts GR00T N1.5~\cite{bjorck2025gr00t} as the baseline.

\subsection{Action Chunking}
Action chunking plays an important role in recent pre-trained VLA models with diffusion-based action heads. At training-time, action chunking enables the policy to learn a sequence of future actions across multiple timesteps. At inference-time, however, action chunks can be executed differently to balance the policy's consistency and reactivity. For example, ACT \cite{zhao2023learning} generates a new action chunk at every timestep and preserves temporal consistency by assembling actions across different timesteps. In GR00T N1.5~\cite{bjorck2025gr00t}, it adopts the same chunk size at training-time and inference-time. In contrast, $\pi_0$ \cite{black2410pi0} uses a larger chunk size (50) during training, but executes smaller chunks (16 or 25) at inference-time, depending on different embodiment setups. SmolVLA \cite{shukor2025smolvla} follows a similar paradigm, referred to as an asynchronous inference mechanism. Recently, DeepConf \cite{fu2025deep} exploits model output confidence as a signal to address the Best-of-N selection problem, but it is limited to large Language Models (LLMs). Similarly, BID \cite{liu2025bidirectional} proposes to select the best action chunk from multiple candidates based on the designed backward coherence and forward contrast scores. However, it is still required to set a fixed chunk size, which limits the reactivity and efficiency of a policy. Complementary to these fixed action chunking methods, in this paper, we investigate the adaptive selection of the optimal action chunk size at inference-time.

\section{Preliminaries}
\label{priliminary}
\subsection{GR00T N1.5}\label{problem formulation}
GR00T N1.5~\cite{bjorck2025gr00t} is a typical Vision-Language-Action foundation model with a flow-matching action head. It consists of two systems, a Vision-Language Model (VLM) to extract vision-language features $\mathbf{\phi}_t$ as condition, and a Diffusion Transformer (DiT) module as the action head $\mathbf{V}_{\mathbf{\theta}}$ for flow-matching. Specifically, given a ground-truth action chunk $\displaystyle \mathbf{A}_t=\{\mathbf{a}_{t}, \mathbf{a}_{t+1}, \cdots, \mathbf{a}_{t+H-1} \} \in \mathbb{R}^{d\times H}$ at timestep $t$, a flow-matching timestep $\displaystyle \tau \in \left[0,1 \right]$ and sampled noise $\mathbf{\epsilon} \sim \mathcal{N} (\mathbf{0}, \mathbf{I})$,  where $d$ denotes the action degrees of freedom and $H$ is the action horizon, the  noised action chunk $\mathbf{A}_t^{(\tau)}$ is computed as $\mathbf A_t^{(\tau)} = \tau \mathbf{A}_t + (1-\tau)\epsilon$. Together with the robot state embedding $\mathbf{q}_t$, the action head prediction $\mathbf{V}_\theta(\phi_t, \mathbf{A}_t^{(\tau)}, \mathbf{q}_t)$ aims to approximate the flow vector field $\displaystyle \epsilon-\mathbf{A}_t$ by minimizing the following loss:
\begin{equation}
\mathcal{L}_{\rm{fm}}(\theta) = \mathbb{E}_\tau\left[ \left\| \mathbf{V}_\theta(\mathbf{\phi}_t, \mathbf{A}_t^{(\tau)}, \mathbf{q}_t) -(\mathbf{\epsilon}-\mathbf{A}_t)   \right\|^2 \right], 
\end{equation}
which is also known as flow-matching loss in popular VLA models. At inference-time, action chunk is generated by $K$-step iterative denoising from random noise $\mathbf{A}_t^{(0)} \sim \mathcal{N}(\mathbf{0}, \mathbf{I})$ with forward Euler integration: $    \mathbf{A}_t^{(\tau+1/K)}=\mathbf{A}_t^{(\tau)} + \frac{1}{K} \mathbf{V}_\theta(\phi_t, \mathbf{A}_t, \mathbf{q}_t)$. For more details, one can refer to the work~\cite{bjorck2025gr00t}. Note that our proposed Adaptive Action Chunking (AAC) algorithm is applicable to all VLA models with a diffusion-based action head. In this study, we use GR00T N1.5~\cite{bjorck2025gr00t} as the baseline of implementation.




\subsection{Action Entropy}
Generally, entropy is computed to quantify the degree of disorder, randomness, or uncertainty. In robotics, the action space may consist of discrete controls (\textit{e.g.}, gripper) or continuous controls (\textit{e.g.}, translation and rotation). Discrete control involves selecting from a finite set of actions, such as choosing a specific gear in a vehicle. Continuous control, on the other hand, involves actions that can take on any value within a certain range, such as adjusting the robot's velocity. Therefore, as described by the works~\cite{niu2022efficient,wangtent,xutest}, for a discrete control $a$, its action entropy can be calculated using the following formula:
\begin{equation}\label{dis}
    E_{\rm{dis}} = -\sum_{a \in \mathcal{A}} p(a)\log(p(a)),
\end{equation}
where $\mathcal{A}$ is the discrete action space and $p(a)$ is the probability of a certain action output from the policy network. For continuous control of $ d$-dimensional actions, we use the Gaussian Differential entropy as follows:
\begin{equation}
    E_t = \frac{1}{2}\log[(2\pi e)^ddet(\Sigma_t)]
\label{con}
\end{equation}
where $det(\Sigma_t)$ is the determinant of the covariance matrix.






\begin{figure*}[t]
\centering
\includegraphics[width=\linewidth]{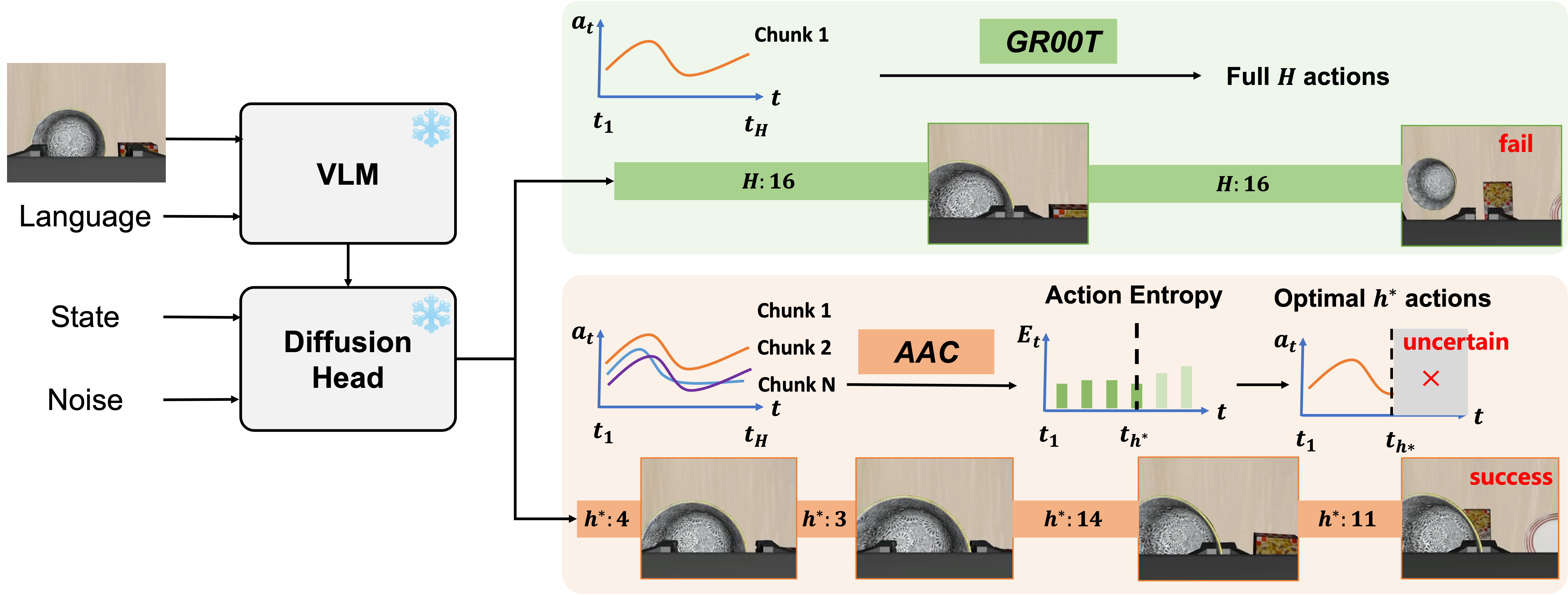}
\caption{\textbf{An overview of AAC}. The proposed Adaptive Action Chunking (AAC) algorithm operates solely at inference-time, without any extra training or architectural changes. Specifically, we exploit the action entropy of continuous and discrete values as the cue to adaptively determine the optimal chunk size $h^*$ in each action chunk at the current observation. Therefore, we can achieve a favorable trade-off between consistency and reactivity in the entire episode, and substantially improve success rates across a variety of manipulation tasks. }

\label{fig:AAC}
\end{figure*}

\section{Adaptive Action Chunking}
\label{method}
Typically, existing VLA models with diffusion-based action heads are trained with an action horizon $H$, and executing a sequence of $h$ ($h\leq H$) actions at each inference timestep~\cite{bjorck2025gr00t,black2410pi0,shukor2025smolvla}. A large fixed chunk size $h$ at inference-time reduces the model's reactivity, whereas a small one compromises its temporal consistency and computational efficiency. Moreover, it is impractical to determine a single fixed chunk size that fits across diverse manipulation tasks. To this end, we seek to adaptively determine the optimal action chunk size $h^*$ to achieve a better balance between consistency and reactivity.

Intuitively, if the model exhibits high uncertainty in its predictions, the action chunk size should be reduced at inference-time. Conversely, when the model is more confident, the action chunk size should be increased. To quantify this uncertainty, we exploit action entropy as a cue to adaptively select the optimal chunk size. Specifically, consider a 7-DOF robotic arm equipped with a gripper as an example, where the predicted action at the $i$-th timestep is denoted as $\mathbf{a}_i$. It consists of three components: 3-DOF translation offsets ($\left[\Delta x, \Delta y, \Delta z\right] \in \mathbb{R}^3 $), 3-DOF rotation offsets ($ \left[\Delta R_x, \Delta R_y, \Delta R_z \right]\in \mathbb{R}^3$) and 1-DOF gripper control $G \in \{0,1\}$. For continuous control of 3-dimensional translations and rotations, the corresponding action entropy $E_t^i$ and $E_r^i$ at the $i$-th timestep are computed according to \cref{con}. For the discrete gripper control, its action entropy $E_g^{i}$ is computed by \cref{dis}. To estimate the probability $p(a)$ in \cref{dis} and the covariance matrix $\Sigma_t$ in \cref{con},  we sample $N$ candidate action chunks in parallel. The probability of predicting a gripper-close state is then estimated as $p(a) = c/N$, where $c$ denotes the number of close-state occurrences among all $N$ samples. Finally, the average action entropy $\overline{E}_h$ with a chunk size $h$ is as follows: 
\begin{equation}
    \overline{E}_h = \frac{1}{h}\sum_{i=t}^{t+h-1} \sum_{j \in \{t,r,g\}} E_j^i.
    \label{average entropy}
\end{equation}
The optimal chunk size $h^*$ in each action chunk $\mathbf{A}_t$ can be adaptively determined by
\begin{equation}
     h^* =\max (\argmax_h(\overline{E}_{h+1}-\overline{E}_h), \ \xi),
    \label{optimal chunk size}
\end{equation}
where $\argmax_h(\overline{E}_{h+1}-\overline{E}_h)$ is the maximum difference point of average action chunk. $\xi$ is a lower bound of chunk size, ensuring a minimum action magnitude to balance temporal consistency and computational efficiency. Since the action spaces of different robot embodiments (e.g., single-arm, bimanual, and humanoid configurations) can be modeled as either continuous or discrete domains, the formulation of action entropy in \cref{average entropy} naturally generalizes across them. For brevity, we omit the detailed derivations in this paper.




\begin{algorithm}
\caption{Adaptive Action Chunking}
\label{chunk size selector}
{\bfseries Input:} $N$ candidate action chunks  $\{\mathbf{A}_t^i\}_{i=1}^N$ predicted in parallel. Each chunk $\mathbf{A}_t^i=\{a_t, a_{t+1}, ..., a_{t+H-1}\}$ includes actions of $H$ future timesteps. 

{\bfseries Output:} Optimal chunk size $h^*$ at current timestep $t$.

\begin{algorithmic}[1]
    \FOR{$i=1$ to $H$}
        \STATE Compute the action entropy of continuous translations and rotations at each timestep $E_t^i$ and $E_r^i$ by \cref{con}.
        \STATE Compute the action entropy of the discrete gripper control at each timestep $E_g^i$ by \cref{dis}.
    \ENDFOR
    
    \FOR{$h=1$ to $H$}
        \STATE Get average action entropy with different chunk size $\overline {E_h}$ by \cref{average entropy}.
    \ENDFOR

    \STATE Find maximum difference point and compute the optimal chunk size $h^*$ by \cref{optimal chunk size}. 
    
    \RETURN Optimal chunk size $h^*$.
\end{algorithmic}
\end{algorithm}


We summarize the above solution as the proposed Adaptive Action Chunking (AAC) algorithm. After selecting the optimal chunk size by AAC, the first $h^*$ actions will be executed at inference-time. For clarity, the whole scheme of our adaptive action chunking is detailed in Algorithm \ref{chunk size selector}. Moreover, the difference between our method and the typical baseline GR00T N1.5~\cite{bjorck2025gr00t} is shown in \cref{fig:AAC}. In consequence, we can achieve a good balance between consistency and reactivity throughout each episode, and substantially improve the success rates on various manipulation tasks.

\section{Experiments}

In this section, to validate the effectiveness of the proposed AAC, we conduct extensive experiments across various manipulation tasks, ranging from simulation benchmarks and real-world applications. 


\begin{table*}[t]
  \caption{\textbf{Main Results on RoboCasa and LIBERO Benchmarks}. We report the success rate (\%) for various subsets and the overall average. Our AAC achieves the best or competitive results across both benchmarks. Bolded entries indicate the highest success rates.}
  \label{tab:main_results}
  \centering
  \small
  
  \setlength{\aboverulesep}{0pt}
  \setlength{\belowrulesep}{0pt}
  \renewcommand{\arraystretch}{1.25} 
  
  \begin{tabularx}{0.95\textwidth}{ l *{4}{C} >{\columncolor[gray]{0.92}}C @{\hspace{1.5em}} *{4}{C} >{\columncolor[gray]{0.92}}C }
    \toprule
    & \multicolumn{5}{c}{\textbf{RoboCasa}} & \multicolumn{5}{c}{\textbf{LIBERO}} \\
    \cmidrule{2-6} \cmidrule{7-11} 
    Method & Relo. & Cont. & Rot. & Button & \textbf{Avg.} & Spatial & Long & Object & Goal & \textbf{Avg.} \\
    \midrule
    GR00T (Default $h$=16) & 42.1 & 80.3 & 57.6 & 80.3 & 59.7 & 93.6 & 88.8 & 97.2 & \textbf{96.8} & 94.1 \\
    \midrule
    GR00T ($h$=2)  & 29.6 & 57.0 & 59.4 & 64.7 & 47.0 & 93.6 & 81.8 & 93.6 & 91.6 & 90.2 \\
    GR00T ($h$=4)  & 37.9 & 71.3 & 61.2 & 78.7 & 56.2 & 94.0 & 85.8 & 95.4 & 95.2 & 92.6 \\
    GR00T ($h$=8)  & 42.8 & 80.3 & \textbf{62.6} & \textbf{81.7} & 61.2 & 92.0 & \textbf{93.4} & 97.8 & 95.4 & 94.7 \\
    GR00T ($h$=12) & 42.9 & 79.5 & 60.0 & 79.3 & 60.2 & 93.0 & 91.0 & 97.6 & 95.2 & 94.2 \\
    \midrule
    GR00T+AAC (\textbf{Ours}) & \textbf{44.4} & \textbf{82.2} & 61.4 & 81.3 & \textbf{62.0} & \textbf{94.4} & 92.8 & \textbf{98.6} & 94.2 & \textbf{95.0} \\
    \bottomrule
  \end{tabularx}
\end{table*}


\subsection{Simulation Experiments}
\subsubsection{Benchmarks}
\noindent \textbf{RoboCasa Kitchen}. RoboCasa \cite{nasiriany2024robocasa} is a simulation benchmark comprising 24 kitchen manipulation tasks. For clarity, 24 RoboCasa tasks are categorized into 4 groups: Relocation (10 pick-and-place tasks), Container (6 tasks of opening or closing doors and drawers), Rotation (5 rotational-affordance tasks such as turning a sink facet or stove knob), and Button (3 tasks involving button pressing on appliances such as a microwave or a coffee machine). Additional details are provided in the supplementary material.

\paragraph{LIBERO.}
LIBERO \cite{liu2023libero} is a robot simulation benchmark that consists of four distinct task suites: LIBERO-Spatial, LIBERO-Object, LIBERO-Goal, and LIBERO-Long. Each suite contains 10 diverse tasks with 50 human-teleoperated demonstrations per task, designed to evaluate a robot’s ability to understand spatial relationships, object interactions, and task-specific
objectives.


\subsubsection{Training and Evaluation Details}
Our model is fine-tuned from the publicly available GR00T N1.5 checkpoint~\cite{bjorck2025gr00t}, which was trained on a large-scale data pyramid. We fine-tune the diffusion head of GR00T N1.5 on eight NVIDIA A800 GPUs with mixed-precision training, while keeping the Eagle-2 VLM \cite{li2025eagle} backbone and vision projector frozen. The updated parameters in the diffusion head comprise DiT backbone, state encoder, state decoder, action encoder, and action decoder. For RoboCasa, we fine-tune 100 demo trajectories generated by MimicGen \cite{mandlekar2023mimicgen} for 90 epochs. For LIBERO, all four task suites (40 tasks in total) are jointly trained for 90 epochs. In this paper, we use the pre-processed dataset provided by OpenVLA \cite{kim2025openvla} and $\pi_0$ \cite{black2410pi0}. We evaluate AAC using 100 rollouts per task in RoboCasa and 50 per task in LIBERO.

\subsubsection{Quantitative Results}
\paragraph{Baseline.} 
As illustrated in \cref{problem formulation}, this paper adopts the vanilla GR00T N1.5~\cite{bjorck2025gr00t} as the baseline, which uses a fixed and consistent action horizon of 16 for both training and action execution. For the compared approaches, inspired by previous studies \cite{chi2023diffusion, black2410pi0, shukor2025smolvla}, we set a fixed inference action chunk size $h$ with different values for GR00T. In contrast, our AAC adaptively determines the optimal chunk size $h^*$ at each inference step.

\begin{figure*}[htbp]
  \centering
  \includegraphics[width=0.8\linewidth]{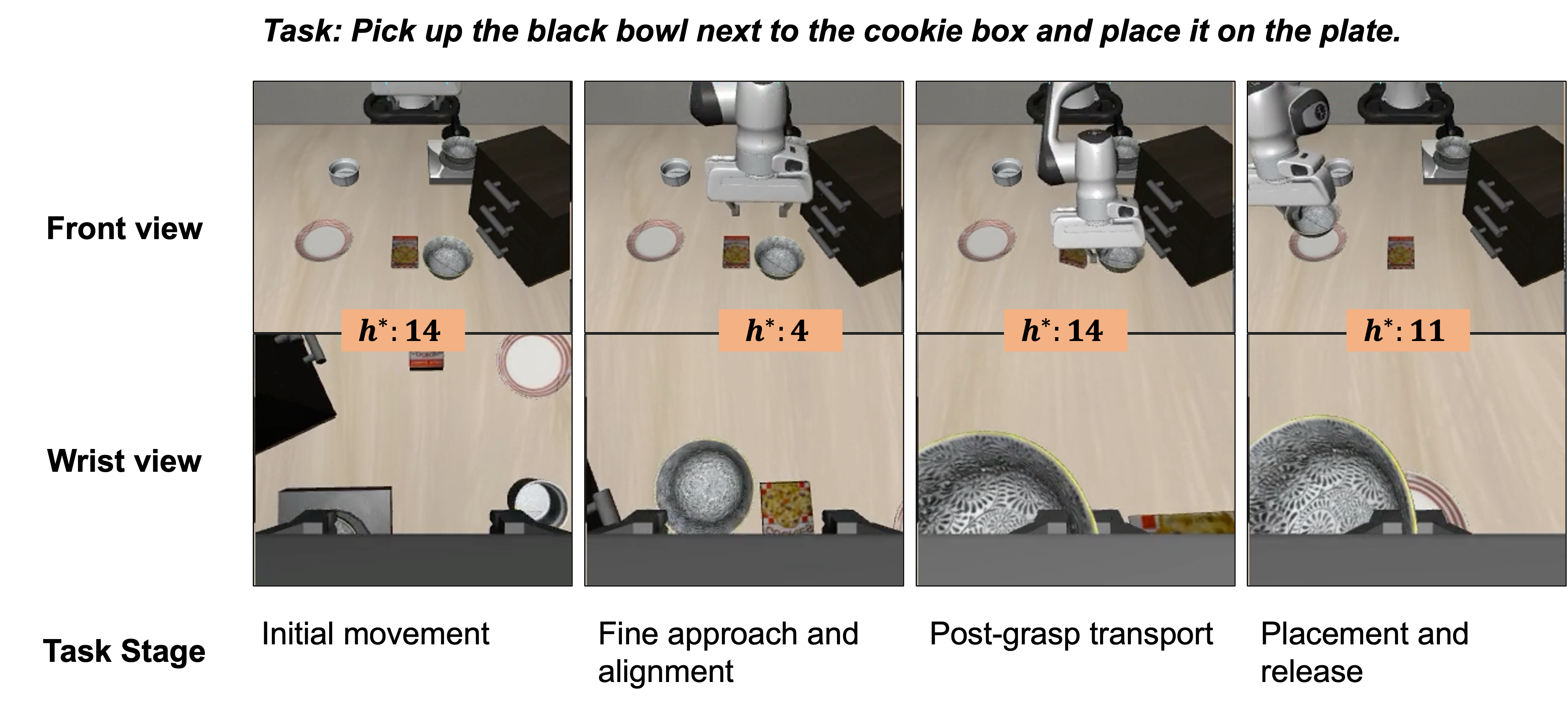}
  \caption{\textbf{Rollout of chunk sizes from AAC}. The derived chunk sizes align with human intuitions with respect to different semantic phases: a large chunk size is observed during the transportation stage, while a small chunk size appears at the critical manipulation stage.}
  \label{fig: demo of chunk size decision}
\end{figure*}

\paragraph{Comparison with Vanilla GR00T.}
Experimental results on the adopted two simulation benchmarks, Robocasa~\cite{nasiriany2024robocasa} and LIBERO~\cite{liu2023libero}, are presented in \cref{tab:main_results}. In Robocasa, AAC improves the task success rate for most tasks, achieving an average gain of $2.3\%$ across 24 tasks. In particular, the success rate on the Rotation task set increases from $57.6\%$ to $61.4\%$, indicating a substantial performance boost. This improvement can be attributed to AAC's ability to handle tasks requiring fine-grained control, such as manipulating a rotational switch on a sink facet or a stove. In contrast, the improvement on the Button task set is relatively modest (from $80.3\%$ to $81.3\%$) \textit{e.g.}, pressing a button on microwaves or coffee machines, since these tasks mainly involve simple pressing actions that demand less precise gripper control. For LIBERO, AAC improves the average task success rate across the four task suites by $0.9\%$ (\textit{i.e.}, from $94.1\%$ to $95.0\%$). The pick-and-place tasks in LIBERO exhibit higher overall success rates compared to those in RoboCasa, likely due to the limited diversity in object positions within the LIBERO testing environments.  Nonetheless, a notable gain of 4\% is achieved on the most challenging LIBERO-Long suite. Each episode in LIBERO-Long involves multiple subtasks requiring precise operation over a longer action trajectory. The proposed adaptive action chunking strategy effectively addresses this challenge by executing high-confidence action chunks and replanning upon new observations, thereby yielding consistent improvements on complex, long-horizon manipulation tasks.



\paragraph{Comparison with Fixed-size GR00T.}
Besides the baseline, vanilla GR00T, we further compare our AAC with GR00T using different chunk sizes. Specifically, we evaluate the task success rates by varying the fixed chunk size $h$ at inference-time. As shown in \cref{tab:main_results}, the results indicate that using a fixed chunk size leads to suboptimal performance. For instance, the task set LIBERO-Spatial performs best with a chunk size of 4, whereas LIBERO-Goal achieves the highest success rate with a chunk size of 16. This observation reveals the potential to improve the task success rate by adaptively changing chunk size within or across different tasks. Inspired by this, AAC adaptively determines the chunk size by leveraging action entropy as a cue, thereby eliminating the need for manual and time-consuming chunk-size tuning. Consequently, AAC consistently outperforms all fixed-size baselines on average. Detailed results for each fixed-size configuration and individual tasks are provided in the Appendix.




\begin{figure}[h]
\centering
\includegraphics[width=\linewidth]{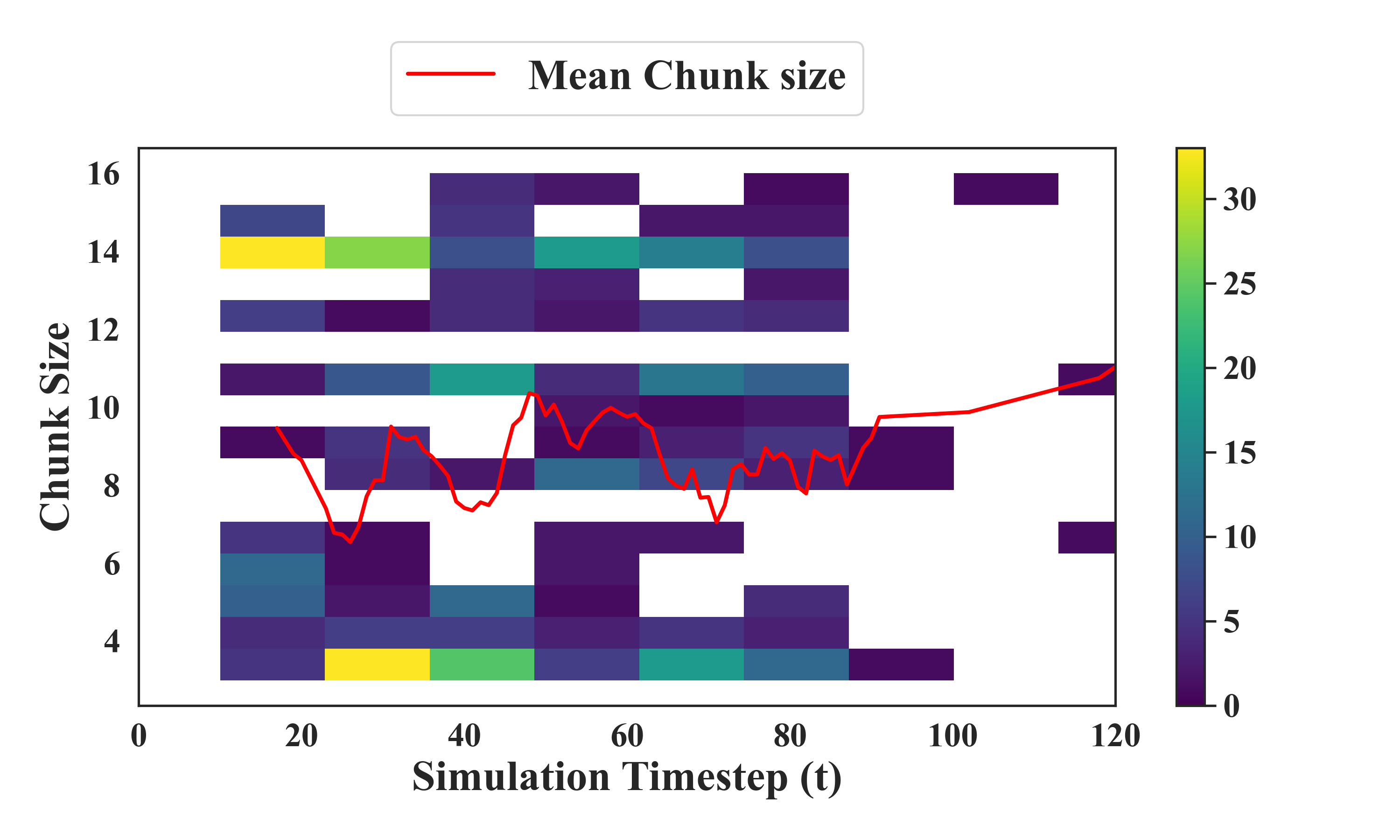}
\caption{\textbf{Distribution of chunk size decisions from AAC}. We show the 
chunk size distribution of episodes on the first task of LIBERO-Spatial: \textit{"Pick up the black bowl next to the cookie box and place it on the plate"}. The heatmap indicates the frequency of different chunk sizes at different decision timesteps. The red curve shows the mean chunk size at different observation timesteps.}
\label{fig:simstep}
\end{figure}

\subsubsection{Qualitative Results}
\paragraph{Distribution of Chunk Size Decision.}

To qualitatively analyze the performance of the proposed AAC, we present a representative example illustrating the chunk size decision process in \cref{fig: demo of chunk size decision}. The corresponding distribution of chunk sizes across simulation timesteps is visualized in \cref{fig:simstep}. As shown in the heatmap and the average curve, the chunk sizes predicted by AAC closely follow the semantic phases of the task. Initially (around $t<20$), when the gripper is far away from the object, AAC tends to predict a large chunk size. When the gripper approaches the object (around $30<t<50$), small chunk sizes are predicted. After successfully grasping the object ($t>50$),  AAC again increases the chunk size to efficiently transport the object to the target location. This adaptive pattern mirrors human intuition: larger chunks enable fast, coarse movements when approaching the target or transporting the lifted object, while smaller chunks with high-frequency observations ensure precise, fine control during the critical grasping and manipulation phases.




\subsubsection{Scalability and Efficiency}

\paragraph{Different Backbones} We apply our AAC to different VLA backbones (\textit{i.e.}, GR00T and $\pi_{0.5}$\footnote{\url{gs://openpi-assets/checkpoints/pi05\_libero/}}\cite{intelligence2025pi}) on LIBERO and LIBERO Pro \cite{zhou2025libero} benchmarks. From the results in \cref{tab: different backbones}, we can observe that AAC generalizes well on $\pi_{0.5}$ backbone. We highlight that AAC is designed for the flow-matching action head, and it can be easily adapted to different backbones.

\begin{table}[t] 
  \caption{Success rates (\%) with $\pi_{0.5}$  backbone on LIBERO.}
  \centering
  \label{tab: different backbones}
  \small 
  \begin{tabular*}{\columnwidth}{@{\extracolsep{\fill}} l ccccc @{}}
    \toprule
    Method & Spatial & Long & Object & Goal & Avg. \\
    \midrule
    $\pi_{0.5}$  & 98.5 & 92.5 & 98.7 & \textbf{98.1} & 97.0 \\ 
    $\pi_{0.5}$+AAC (\textbf{Ours}) & \textbf{99.1} & \textbf{95.2} & \textbf{99.2} & 98.0 & \textbf{97.9} \\
    \bottomrule
  \end{tabular*}
\end{table}


\paragraph{Out of Distribution Scenarios} We evaluate the robustness of AAC on OOD tasks by using the position perturbation subset of LIBERO-Pro \cite{zhou2025libero} benchmark. The quantitative results are reported in \cref{tab:libero-pro}, from which we can summarize that as the perturbation level increases, the success rates of both the baseline methods and AAC decrease. Nevertheless, our AAC demonstrates its effectiveness by leveraging the action entropy to dynamically modulate the execution horizon.

\begin{table}[t]
  \caption{Success rates (\%) on LIBERO-Pro~\cite{zhou2025libero}.}
  \label{tab:libero-pro}
  \centering
  \small
  \begin{tabular*}{\columnwidth}{@{\extracolsep{\fill}} l cccc @{}}
    \toprule
    Perturbation level   & $\times$0.2 & $\times$0.3 & $\times$0.4 & Avg. \\
    \midrule
    GR00T         & 11.3 & 0.4  & 0.0  & 3.9  \\
    GR00T+AAC (\textbf{Ours}) & \textbf{17.9} & \textbf{0.9}  & 0.0  & \textbf{6.3}  \\
    \midrule 
    $\pi_{0.5}$   & 53.2 & 29.9 & 9.5  & 30.9 \\
    $\pi_{0.5}$+AAC (\textbf{Ours}) & \textbf{57.4} & \textbf{35.3} & \textbf{11.8} & \textbf{34.8} \\
    \bottomrule
  \end{tabular*}
\end{table}


\paragraph{Different Number of Samples.} As illustrated in \cref{method}, AAC is based on the distribution of action entropy, which is derived from multiple candidate predicted action chunks. Therefore, we investigate the impact of varying the number of samples. From the results in \cref{tab: different number of samples}, we can conclude that the success rate can be continuously improved with a larger number of samples. When the number of samples increases from 1 to 20, the success rate continuously improves $0.9\%$, from $94.1\%$ to $95.0\%$. However, the improvement is marginal when the samples are enough (\textit{e.g}, 20) to estimate the action entropy. In consequence, we use 20 samples to balance the tradeoff between computational efficiency and success rate in all benchmarks. 


\paragraph{Computational Efficiency.}
The calculation of action entropy requires multiple action chunk candidates. Thus, AAC inevitably involves additional computational cost compared to the fixed-size competitors. Fortunately, as the multiple action chunk candidates can be generated through batched parallel computation, the computational cost is negligible.  The inference time for different sample sizes on a single A800 GPU is shown in \cref{tab: different number of samples}. From the results, we can see that the influence on the inference speed is negligible when the number of samples is smaller than 10. The default sample size used in this paper is 20, which will introduce about 20 ms inference delay according to the testing results in the last row of \cref{tab: different number of samples}. Higher-end GPUs can further close the computational gap.

\begin{table}[t]
  \caption{Success rates (\%) on LIBERO and inference-time under different numbers of samples for estimating action entropy.}
  \label{tab: different number of samples}
  \centering
  \begin{tabular}{@{}lcccccc@{}}
    \toprule
    \#Samples   & 1     & 5     & 10    & 20       & 30       & 40   \\
    \midrule
    Succ. (\%)  & 94.1   & 94.7  & 94.4  & 95.0     & 95.0     & 95.5 \\
    Time (ms)     & 83.0    & 83.5   & 84.3   & 106.0      & 136.5      & 157.0 \\
    \bottomrule
  \end{tabular}
\end{table}






\begin{figure*}[htbp]
  \centering
  \includegraphics[width=0.9\linewidth]{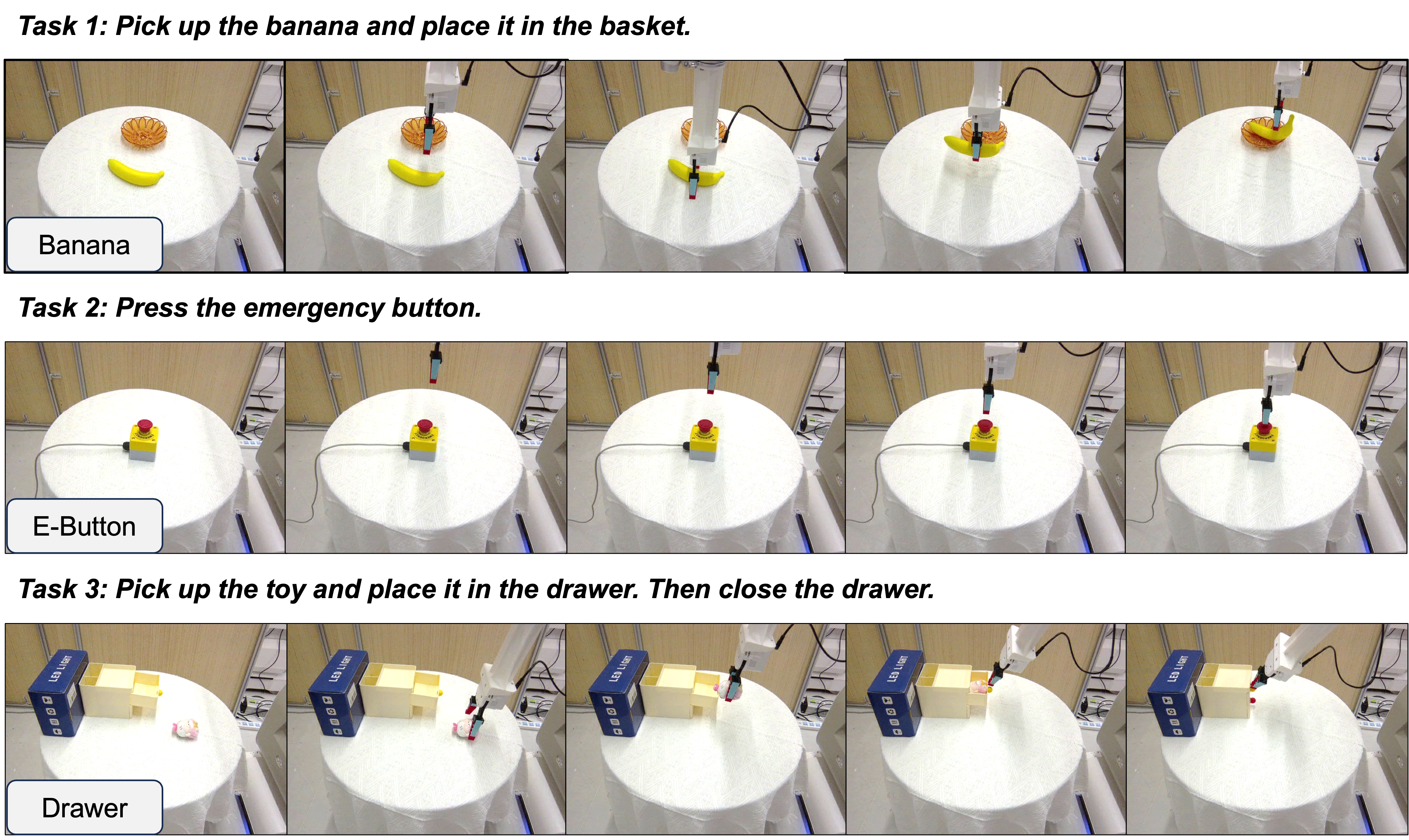}
  \caption{Execution examples for real-world tasks using AAC. Videos of complete execution trajectories will be publicly available.}
  \label{fig: real world tasks}
\end{figure*}

\subsection{Real-world Experiments}
\subsubsection{Setup}
We further evaluate the performance of the proposed AAC with different desktop applications in the real world. We use a Realman single-arm robot with a Mycobot gripper. A SpaceMouse is also used for collecting 50 human-teleoperated demonstrations per task. The visual observation includes a wrist RGB view and a side RGB view on a tripod. The state and action space are consistent with the configuration in Robocasa and LIBERO benchmarks. We finetune GR00T N1.5~\cite{bjorck2025gr00t} checkpoint on each of the tasks with 8 NVIDIA A800 GPUs for 300 epochs. The batch size on each GPU is set to 32. For each task, we evaluate the success rate on 20 trials with a limitation of 600 action execution steps for each trial.


\subsubsection{Tasks} We design 3 real-world tasks to evaluate the proposed method: (1) a banana pick-and-place task, (2) an emergency button pressing task, and (3) a long-horizon task that involves picking and placing the toy, then closing the drawer. Specifically, the first task requires picking up a banana and placing it into a rattan fruit basket. Both the banana and the basket are randomly positioned on a round table with a diameter of 0.6 m. For evaluation, 20 different combinations of banana and basket locations are tested, as the position and orientation of the gripper critically influence grasping success. The second task involves pressing an emergency button (E-Button), demanding higher localization accuracy than the banana pick-and-place task. During data collection, the E-Button positions cover only the upper half of the table. Therefore, this task is challenging because some button positions in the evaluation are unseen during training. The third task is a long-horizon task, where the robot must pick up a randomly placed toy, place it into a drawer, and subsequently close the drawer. To complete this task, the policy needs to ensure precision along the whole trajectory.



\begin{table}[h]
  \caption{Success rates (\%) on real-world applications.}
  \label{tab: real-world results}
  \centering
  \begin{tabular}{@{}lcccccc@{}}
    \toprule
    Task     & Banana  & Button & Drawer   & Avg.  \\
    \midrule
    GR00T    &  70.0        &   65.0        &  65.0                    & 67.0  \\
    GR00T+AAC & \textbf{90.0} & \textbf{75.0}  & \textbf{80.0}    &  \textbf{82.0} \\
    \bottomrule
  \end{tabular}
\end{table}

\subsubsection{Results}
\paragraph{AAC improves real-world task success rates.}
We present the quantitative results in \cref{tab: real-world results}. The results demonstrate that AAC consistently outperforms the baseline GR00T across all three real-world tasks. On average, AAC achieves a $15\%$ performance gain, increasing the success rate from $67.0\%$ to $82.0\%$. The performance improvement arises from multiple aspects. For the banana pick-and-place task, AAC allows the robot to better align the gripper orientation with respect to the target banana, resulting in more precise grasping. In the E-Button pressing task, both the baseline GR00T and the proposed AAC can successfully complete the task under familiar conditions. However, when the E-button appears at locations that lie near but outside the location distribution of human-teleoperated demonstrations, AAC shows better generalization ability and achieves more robust performance. For the long-horizon toy manipulation task, although the drawer-closing stage is relatively easy due to its lower precision requirements, AAC achieves a higher success rate in the toy-picking stage, benefiting from adaptive chunk sizes.



\begin{figure}[h]
  \begin{subfigure}{0.23\textwidth}
    \includegraphics[width=\linewidth]{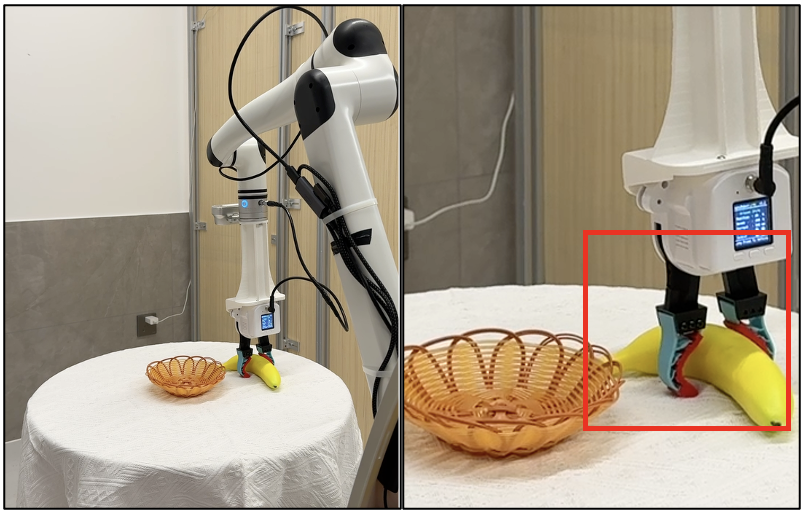}
    \caption{Execution process of GR00T.}
  \end{subfigure} 
  \begin{subfigure}{0.23\textwidth}
    \includegraphics[width=\linewidth]{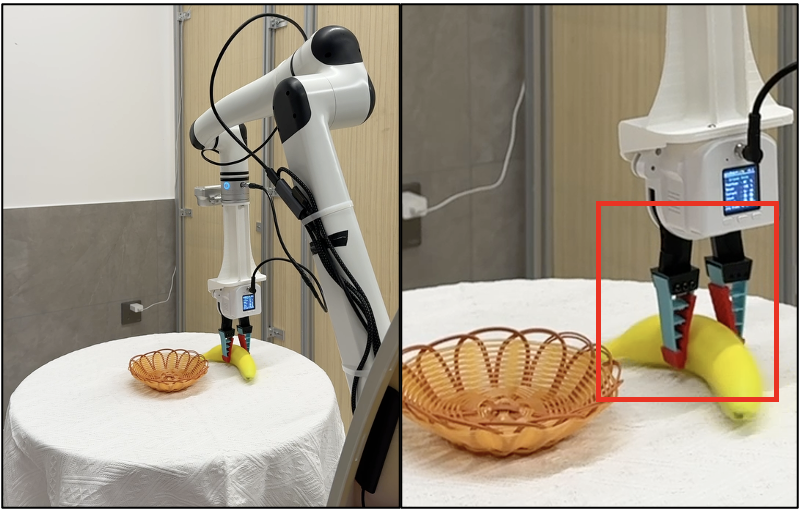}
    \caption{Execution process of AAC.}
  \end{subfigure} 
  \caption{\textbf{AAC improves action accuracy and safety}. \textbf{Left}: the gripper collided with the table. \textbf{Right}: the gripper reached an appropriate lowest point.}
  \label{fig: banana comparison}
\end{figure}

\paragraph{AAC improves action accuracy and safety.}
In this part, we illustrate that the proposed AAC has the ability to localize more precisely, thanks to the adaptive chunk size strategy that ensures higher action quality. As shown in \cref{fig: banana comparison}, the baseline vanilla GR00T tends to collide with the tabletop due to the low-quality actions predicted at the earlier observation point. In contrast, AAC is able to smoothly stop at an appropriate lowest point by filtering high-entropy or uncertain actions through the dynamically selected chunk size. This phenomenon demonstrates the potential of AAC for high-precision robot manipulation tasks, thus improving safety and robustness in practice.

\section{Conclusion}
In this work, we have analyzed that the action chunk size is important for robot manipulation tasks.  Based on this observation, we propose Adaptive Action Chunking (AAC), an inference-time strategy that exploits action entropy as the cue to adaptively select action chunk size. It aligns with human intuitions that a large chunk size is used at the transportation stage, while a small chunk size appears at the critical manipulation stage. Consequently, we achieve a good balance between reactivity and consistency.  Extensive experiments on a variety of simulation benchmarks and real-world applications have validated the effectiveness of our approach over the baseline and state-of-the-art alternatives.

{
    \small
    \bibliographystyle{ieeenat_fullname}
    \bibliography{main}
}

\clearpage
\setcounter{page}{1}
\maketitlesupplementary

\appendix

\section{Minimum Action Magnitude}

To further guarantee temporal consistency and computational efficiency, we impose a minimum action magnitude constraint on the optimal action chunk size, thereby defining a dynamic threshold $\xi$ in \cref{minimum id of action magnitude}:

\begin{equation}
\label{minimum id of action magnitude}
    \xi = \argmin_l(m(l) > \alpha)
\end{equation}
where the action magnitude $m(l)$ is defined as the sum of the component magnitudes of translation $m_t$, rotation $m_r$, and gripper motion $m_g$, as illustrated in \cref{action magnitude total}. The parameter $\alpha$ denotes the minimum movement energy of the robot arm, which we set to 3 in our experiments.

\begin{equation}
    \label{action magnitude total}
     m(l) = \sum_{i \in \{t,r,g\}} m_i(l)
\end{equation}



Specifically, the translational action magnitude is defined as the Euclidean norm of the translation vector from the start point of an action chunk to the end point of a selected chunk size $l$, as illustrated in \cref{action magnitude position}. The rotational action magnitude is obtained by accumulating the rotational offsets through sequential quaternion composition, as shown in \cref{action magnitude rotation}. For the gripper action magnitude, we set it to $1$ when the gripper state changes (e.g., from open to closed or vice versa), and to $0$ otherwise, as presented in \cref{action magnitude gripper}. Here, $\mathbf{1}$ denotes an indicator function that returns $1$ if the condition is satisfied and $0$ otherwise.

\begin{equation}
    \label{action magnitude position}
    m_{t}(l) = \left\|\sum_{l} \Delta x, \sum_l \Delta y, \sum_{l} \Delta z) \right\|
\end{equation}

\begin{equation}
    \label{action magnitude rotation}
    m_{r}(l) = \left\| \prod_{l} \Delta q \right\|
\end{equation}

\begin{equation}
    \label{action magnitude gripper}
    m_g(l) = \mathbf{1}_\mathrm{switch}
\end{equation}

\section{Complete Experiment Results in RoboCasa}

The complete experimental results for all 24 tasks in RoboCasa are presented in \cref{tab: RoboCasa baseline full} and \cref{tab: fixed_size results robocasa full}. On average, AAC demonstrates the best overall performance across these tasks. Nevertheless, there remains room for improvement in optimizing the chunk size selection for future work, as the optimal performance for each task has not yet been fully achieved.

\begin{table}[tb]
  \caption{Success rates (\%) on RoboCasa.}
  \label{tab: RoboCasa baseline full}
  \centering
  \small
  \setlength{\tabcolsep}{3pt} 
  
  \begin{tabularx}{\columnwidth}{@{} l >{\raggedright\arraybackslash}X c c @{}}
    \toprule
    Task Group & Task Name & GR00T  & GR00T + AAC \\ 
    \midrule
    \multirow{10}{*}{Relocation} 
    & Counter to Cab & 57.0 & 59.0\\ 
    & Counter to Stove & 41.0 & 40.0\\ 
    & Sink to Counter & 46.0 & 52.0\\ 
    & Stove to Counter & 60.0 & 63.0\\ 
    & Counter to Sink & 26.0 & 28.0\\ 
    & Counter to Microwave & 33.0 & 30.0\\ 
    & Cab to Counter & 26.0 & 25.0\\ 
    & Microwave to Counter & 27.0 & 36.0\\ 
    & Coffee Setup Mug & 38.0 & 39.0\\ 
    & Coffee Serve Mug & 67.0 & 72.0\\ 
    \midrule 
    \multirow{6}{*}{Container} 
    & Open Single Door & 62.0 & 73.0\\ 
    & Close Single Door & 94.0 & 92.0\\ 
    & Close Double Door & 56.0 & 62.0\\ 
    & Open Double Door & 89.0 & 88.0\\ 
    & Open Drawer & 81.0 & 80.0\\ 
    & Close Drawer & 100.0 & 98.0\\ 
    \midrule
    \multirow{5}{*}{Rotation} 
    & Turn Sink Spout & 69.0 & 66.0\\ 
    & Turn on Sink facet & 85.0 & 91.0\\ 
    & Turn off Sink Facet & 74.0 & 80.0\\ 
    & Turn on Stove & 39.0 & 49.0\\ 
    & Turn off Stove & 21.0 & 21.0\\ 
    \midrule
    \multirow{3}{*}{Button} 
    & Coffee Press Button  & 94.0 & 93.0\\ 
    & Turn on Microwave & 66.0 & 68.0\\ 
    & Turn off Microwave & 81.0 & 83.0\\ 
    \midrule
    & Relocation Average & 42.1 & 44.4\\
    & Container Average & 80.3 & 82.2\\
    & Rotation Average & 57.6 & 61.4\\
    & Button Average & 80.3 & 81.3\\
    \rowcolor[gray]{0.92}
    & 24 Tasks Average & 59.7 & 62.0\\
    \bottomrule
  \end{tabularx}
\end{table}

\begin{table*}[tb]
  \caption{Success rates (\%) with Different Chunk Sizes.}
  \label{tab: fixed_size results robocasa full}
  \centering
  \small
  \begin{tabular}{@{}llccccc@{}}
    \toprule
    Task Group & Task Name & Fixed-2    & Fixed-4     & Fixed-8   & Fixed-12    & Fixed-16 \\
    \midrule
    \multirow{8}{*}{Relocation}
    & Counter to Cab & 51.0 & 68.0 & 61.0 & 57.0 & 57.0 \\
    & Counter to Stove & 31.0 & 40.0 & 46.0 & 45.0 & 41.0 \\
    & Sink to Counter & 35.0 & 44.0 & 56.0 & 50.0 & 46.0 \\
    & Stove to Counter & 59.0 & 62.0 & 69.0 & 64.0 & 60.0 \\
    & Counter to Sink & 6.0 & 10.0 & 15.0 & 19.0 & 26.0 \\
    & Counter to Microwave & 7.0 & 12.0 & 29.0 & 23.0 & 33.0 \\
    & Cab to Counter & 18.0 & 18.0 & 23.0 & 23.0 & 26.0 \\
    & Microwave to Counter & 27.0 & 42.0 & 28.0 & 36.0 & 27.0 \\
    & Coffee Setup Mug & 31.0 & 39.0 & 40.0 & 46.0 & 38.0 \\
    & Coffee Serve Mug & 31.0 & 44.0 & 61.0 & 66.0 & 67.0 \\
    \hline
    \multirow{6}{*}{Container} 
    & Open Single Door & 15.0 & 41.0 & 61.0 & 62.0 & 62.0 \\
    & Close Single Door & 84.0 & 89.0 & 87.0 & 95.0 & 94.0 \\
    & Close Double Door & 53.0 & 62.0 & 73.0 & 58.0 & 56.0 \\
    & Open Double Door & 44.0 & 73.0 & 81.0 & 84.0 & 89.0 \\
    & Open Drawer & 51.0 & 64.0 & 81.0 & 79.0 & 81.0 \\
    & Close Drawer & 95.0 & 99.0 & 99.0 & 99.0 & 100.0 \\
    \hline
    \multirow{5}{*}{Rotation} 
    & Turn Sink Spout & 70.0 & 69.0 & 69.0 & 67.0 & 69.0 \\
    & Turn on Sink facet & 84.0 & 87.0 & 85.0 & 83.0 & 85.0 \\
    & Turn off Sink Facet & 82.0 & 78.0 & 85.0 & 80.0 & 74.0 \\
    & Turn on Stove & 42.0 & 52.0 & 52.0 & 47.0 & 39.0 \\
    & Turn off Stove & 19.0 & 20.0 & 22.0 & 23.0 & 21.0 \\
    \hline
    \multirow{3}{*}{Button} 
    & Coffee Press Button  & 76.0 & 89.0 & 98.0 & 98.0 & 94.0 \\
    & Turn on Microwave & 47.0 & 51.0 & 61.0 & 60.0 & 66.0 \\
    & Turn off Microwave & 71.0 & 96.0 & 86.0 & 80.0 & 81.0 \\
    \hline
    & Relocation Average & 29.6 & 37.9 & 42.8 & 42.9 & 42.1 \\
    & Container Average & 57.0 & 71.3 & 80.3 & 79.5 & 80.3 \\
    & Rotation Average & 59.4 & 61.2 & 62.6 & 60.0 & 57.6 \\
    & Button Average & 64.7 & 78.7 & 81.7 & 79.3 & 80.3 \\
    \rowcolor[gray]{0.92}
    & 24 Tasks Average & 47.0 & 56.2 & 61.2 & 60.2 & 59.7 \\ 
    \bottomrule
  \end{tabular}
\end{table*}


\end{document}